# Improving Semiconductor Device Modeling for Electronic Design Automation by Machine Learning Techniques

Zeheng Wang, Member, IEEE, Liang Li, Ross C. C. Leon, Jinlin Yang, Junjie Shi, Timothy van der Laan, and Muhammad Usman

*Abstract*— The semiconductors industry benefits greatly from the integration of Machine Learning (ML)-based techniques in Technology Computer-Aided Design (TCAD) methods. The performance of ML models however relies heavily on the quality and quantity of training datasets. They can be particularly difficult to obtain in the semiconductor industry due to the complexity and expense of the device fabrication. In this paper, we propose a self-augmentation strategy for improving ML-based device modeling using variational autoencoder-based techniques. These techniques require a small number of experimental data points and does not rely on TCAD tools. To demonstrate the effectiveness of our approach, we apply it to a deep neural network-based prediction task for the Ohmic resistance value in Gallium Nitride devices. A 70% reduction in mean absolute error when predicting experimental results is achieved. The inherent flexibility of our approach allows easy adaptation to various tasks, thus making it highly relevant to many applications of semiconductor industry.

Zeheng Wang and Muhammad Usman are with Data61, CSIRO, Clayton, 3168, VIC, Australia. (e-mail: zenwang@outlook.com, muhammad.usman@csiro.au).

Zeheng Wang is also with CSIRO Manufacturing, P. O. Box 218, 36 Bradfield Road, Lindfield, NSW 2070, Australia.

Liang Li is with the Academy for Advanced Interdisciplinary Studies, Peking University, Beijing 100871, China.

Ross C. C. Leon is with Quantum Motion, 9 Sterling Way, London N7 9HJ, United Kingdom.

Jinlin Yang is with Department of Chemistry, National University of Singapore, 3 Science Drive 3, 117543, Singapore.

Junjie Shi is with School of Materials Science and Engineering, University of New South Wales, Sydney, NSW 2052, Australia

Timothy van der Laan is with CSIRO Manufacturing, P. O. Box 218, 36 Bradfield Road, Lindfield, NSW 2070, Australia. (e-mail: tim.vanderlaan@csiro.au)

Corresponding author: Zeheng Wang, Timothy van der Laan, and Muhammad Usman.

## I. Introduction

Electronic Design Automation (EDA) has been crucial in advancing the semiconductors industry by simplifying design tasks and reducing their time consumption [1]. One particular EDA technique, Technology Computer-Aided Design (TCAD), has been especially useful in the area of semiconductor devices. TCAD solves basic physics equations using the finite element method, such as the Poisson and Schrödinger equations, which provides easy access to simulated results that would be difficult to solve manually [2]–[4]. Additionally, TCAD has significantly reduced the cost of experiments during device design by avoiding them altogether [5].

Nevertheless, simulating complex three-dimensional device structures requires significant computational resources. While many models and methods have been developed to reduce resource consumption, exploring novel methodologies of TCAD remains a pressing issue to balance the accuracy and time consumption of sophisticated physics simulations. So far, Machine Learning (ML)-based solutions have been successfully employed in many device modeling cases and offer the advantage of low-resource consumption after model training [6]–[9]. However, with expanding size of the ML models, there is an increasing need for input data to fully complete model training [10].

TCAD-based data augmentation, a technique that has garnered significant attention in the semiconductor industry since 2019 [9], [11], [12], has been employed to generate artificial data that can be fed into Deep Neural Network (DNN)-based models. This approach could provide an expanded dataset and then significant boost to DNN-based modeling within the TCAD industry's development. However, many problems in the semiconductor industry cannot be directly solved by TCAD tools, such as the simulation of the formation of ohmic contacts in Gallium Nitride (GaN) devices, which imposes a formidable challenge on the TCAD-based augmentation technique.

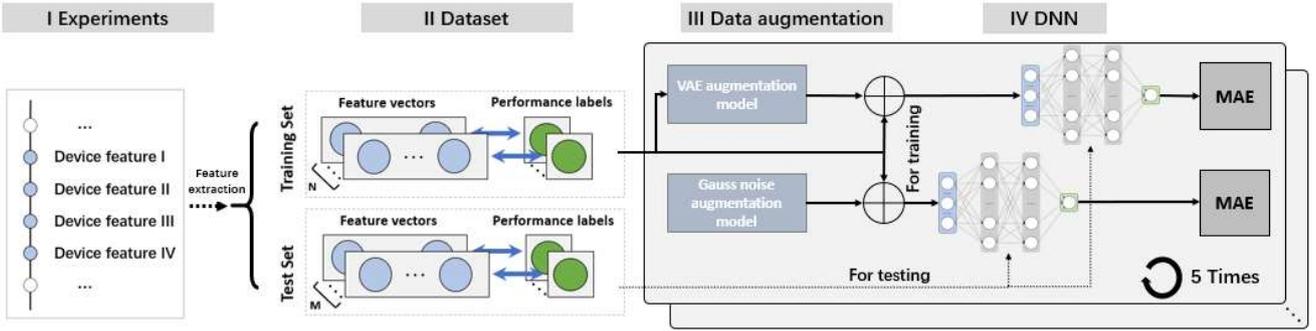

Fig. 1. The sketch of the whole procedure of augmentation-enhanced ML-based semiconductor device modeling – The data was extracted from literature containing experimental data, and then was augmented by the augmentation model (VAE). The original data and the artificially augmented data were measured in a DNN-based prediction task. The data flows between each step are indicated by arrows. Note that the validation process was carried out with the experimental data that were extracted from the recently reported literature.

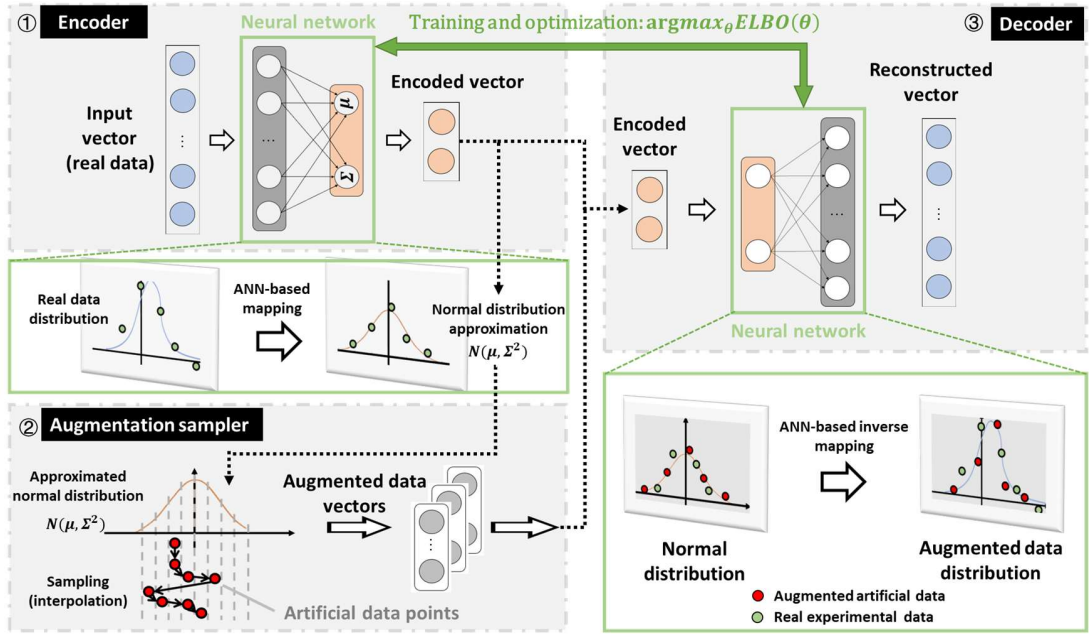

Fig. 2. The schematic procedure of artificial data generation. The typical VAE model was adopted for generating the features, while for generating the corresponding labels, the nearest neighbor algorithm was used. The green frames indicate the process of data transformation in latent spaces, while the frames with grey backgrounds show the structure of the augmentation model.

This paper proposes a novel data self-augmentation strategy for expanding the size of semiconductor device datasets used for Deep Neural Network (DNN)-based modeling tasks, without requiring calibration of TCAD tools. Our approach is based on the use of Variational Auto-Encoder (VAE) [13], a type of generative model that can learn the underlying probability distribution of the dataset, and then generate new, synthetic data samples that are statistically similar to the original data. Specifically, we apply the data from our proposed strategy to the DNN-based modeling task -- predicting the ohmic contacts of Gallium Nitride (GaN) devices, a challenging problem that cannot be solved directly using TCAD.

To validate the effectiveness of our approach, we used experimental data extracted from the literature to train the VAE, which was then used to generate augmented data that was combined with the experimental data for training the DNN model (for dataset details see our previous work [14]). The results demonstrate that our data augmentation strategy significantly reduces the mean absolute error of the prediction by up to 70% for AlGaN/GaN devices, when compared with the model using experimental data only. This finding highlights the potential of our approach for enhancing the accuracy and robustness of device simulation tools in the semiconductor industry.

## II. METHODS FOR DATA AUGMENTATION AND VERIFICATION

In this paper, we propose a data self-augmentation strategy based on a Variational Autoencoder (VAE) to improve the performance of ML-based device modeling. Generally, ML-based device modeling aims to link device features, such as gate length, drain voltage, and annealing temperature, to performance, such as surface potential, saturation current, and on/off ratio [15]–[17]. Our proposed data augmentation framework is divided into three main parts: a feature preprocessing module (sectors I and II), a data augmentation module (sector III), and an ML-based modeling module (sector IV), as shown in Fig. 1. The first step of modeling is to extract parameter data, i.e., device features, from experimental results, as illustrated in sector I. The extracted data is then transformed into vectors by the feature preprocessing module, as shown in sector II. The experimental data is split into a training set (67% of the data) and a test set (33% of the data). The training set is used to build the generative models for augmentation, as depicted in sector III. The generated artificial data is combined with the training set and fed into the deep neural network (DNN)-based model for semiconductor device performance modeling (sector IV). The test set is used to evaluate the generalization ability of the DNN and the augmentation model.

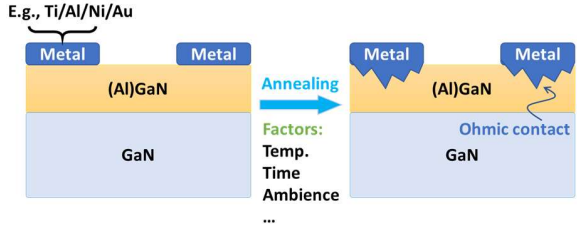

Fig. 3. The typical Ohmic contact structure of GaN device and the simplified fabrication process flow.

The steps of the proposed data augmentation strategy and verification are described as follows, with technical details provided in the subsequent subsections:

1) *Extract feature data from experiments and split the data into a test set and a training set.*

2) *Train the VAE-based model using the training set.*

3) *Generate artificial data and combine it with the training set.*

4) *Train the corresponding DNN models using the combined dataset.*

5) *Test the DNN model using the test set and calculate the Mean Absolute Error (MAE).*

6) *Repeat steps 2) - 5) five times and calculate the average MAE for further discussion.*

### A. Feature pre-processing

To minimize the influence of missing data points, all experimental data was filtered by a 'data cleaning' process, this deletes whole vectors if any vacancy exists. Then the experimental data was divided into two parts for the next step: the numerical part and the text-based part. These two parts are handled by different processes: for the text-based input, such as the name of material layers, the data was then transformed into a numerical vector using one-hot encoding [14], [18]. For the numerical input, the item x of the data was directly standardized by z-score to ensure the numerical input was centered to 0 with a standard deviation of 1, following the equation below:

$$z = \frac{x - \mu_x}{\sigma_x} \qquad (1)$$

where $\mu_x$ is mean of $x$ and $\sigma_x$ is the standard deviation of $x$. Note, for simplicity the ohmic value data described in the following sections are all standardized values, not original values.

### B. Generative model

The primary goal of a generative model is to learn the joint probability distribution of a given dataset through unsupervised learning. This enables successful data augmentation from experimental data by interpolating variations into the trained generative model (shown in panels I and II of Fig. 2). In this study, we used a two-component generator consisting of an artificial feature generator and an artificial label generator. To overcome the challenge of insufficient training data in the augmentation task for the artificial feature generator, we utilized the VAE, which is a simple yet powerful deep generative model. Additionally, we applied a K-th Nearest Neighbor (KNN) regressor to generate corresponding labels. This approach offers an effective solution to the problem of insufficient training data in data augmentation and represents a significant contribution to the field of generative modeling.

#### 1) Artificial Feature Generator

The artificial feature generator is realized by the VAE. This is a variant of the automatic encoder combining variational inference with a conventional autoencoder framework. Thus, the VAE consists of two parts: an encoder and decoder. The encoding-decoding process efficiently realizes dimensionality reduction, which emphasizes the preferred features of the input and suppresses the less-important features to minimize interference [19], [20]. The encoder encodes, denoted as $Enc(x)$, input $x$ into a latent representation $z$ with a parameter $\theta$, which is denoted by $q_\theta(z|x)$. The decoder reconstructs, denoted as $Dec(z)$, the data distribution $\tilde{x}$ from the given $z$, which is depicted as follows:

$$z = Enc(x) \sim q_\theta(z|x) \qquad (2)$$
$$\tilde{x} = Dec(z) \sim p(x|z) \qquad (3)$$

Given a dataset $x = \{x_1 \dots, x_N\}$, where $N$ is the number of samples, the target of the generative model is to maximize the probability $p(X)$:

$$p(X) = \sum_{i=1}^{N} p(x_i|z) \, p(z) \qquad (4)$$

in which $p(z)$ is the probability distribution of the encoded latent representations, which is unknown. The purpose of the VAE is to infer $p(z)$ from the ideal posterior probability $p(z|x)$. It could be replaced by a simpler normal distribution $q_\theta(z|x)$. Then, the problem is converted into minimizing the difference between those two distributions using KL divergence [21]:

$$KL(q_\theta(z|x)||p(z|x))$$
$$= \mathbb{E}[\log q_\theta(z|x) - \log p(z|x) - \log p(z)] \qquad (5)$$

By applying the following function named Evidence Lower Bound (ELBO) [22],

$$ELBO(\theta) = -\mathbb{E}[\log p(x|z) + \log p(z) - \log q_\theta(z|x)]$$
$$= \mathbb{E}[\log p(x|z)] - KL(q_\theta(z|x)||p(z)) \qquad (6)$$

the Eq. (4) can be rewritten as a log likelihood function:

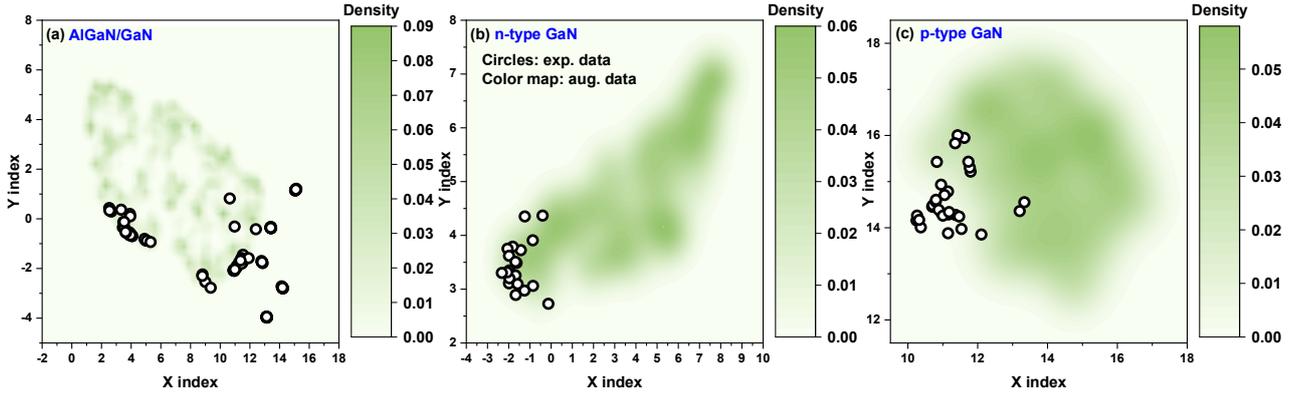

Fig. 4. Visualization of experimental data and augmented data, projected to a two-dimensional plane through a UMAP algorithm, for the resistance value of Ohmic contacts on (a) AlGaN/GaN heterojunction, (b) n-type GaN and (c) p-type GaN substrates. All data for training the VAE were extracted from experiments (for dataset details see our previous work [14]).

$$\log p(\boldsymbol{x}) = KL(q_{\theta}(\boldsymbol{z}|\boldsymbol{x})||p(\boldsymbol{z}|\boldsymbol{x})) + ELBO(\theta) \quad (7)$$

Note that $\boldsymbol{x}$ is known, $\log p(\boldsymbol{x})$ is constant. Besides, KL divergence is always greater than or equal to zero according to Jensen's inequality [23]. Therefore, minimizing the KL divergence is equivalent to maximizing $ELBO(\theta)$. Since no datapoint shares its latent $\boldsymbol{z}$ with another datapoint in VAE, we can write this function for a single datapoint as:

$$ELBO(\theta)_i = \mathbb{E}[\log p(\boldsymbol{x}_i|\boldsymbol{z}_i)] - KL(q_{\theta}(\boldsymbol{z}_i|\boldsymbol{x}_i)||p(\boldsymbol{z}_i)) \quad (8)$$

Thus, the training target of the VAE is actually approaching the maximum of the ELBO function shown below, which is also labeled in Fig. 2 between the panel I and III:

$$argmax_{\theta}ELBO(\theta) = \sum_i ELBO(\theta)_i \quad (9)$$

According to [20], $p(\boldsymbol{z}_i) = \mathcal{N}(0, I)$ is assumed as a standard normal distribution, while $q_{\theta}(\boldsymbol{z}_i|\boldsymbol{x}_i) = \mathcal{N}(\mu(\boldsymbol{x}_i), \Sigma(\boldsymbol{x}_i))$, where $\mathcal{N}$ represents a normal distribution, $I$ is an identity matrix, and, $\mu$ and $\Sigma$ are arbitrary deterministic functions that can be learned from data. Thus, this can be explicitly expressed by $\mu$ and $\Sigma$ as follows:

$$\sum_{i,k}\left[\left(x_i^k - \tilde{x}_i^k\right)^2 + \frac{1}{2}\left(\Sigma(\boldsymbol{x}_i)_k + \mu^2(\boldsymbol{x}_i)_k - 1 - \log\Sigma(\boldsymbol{x}_i)_k\right)\right] \quad (10)$$

where $\tilde{x}_i^k$ is the $k$-th element of reconstructed data vectors $\tilde{\boldsymbol{x}}_i$, the $\mu(\boldsymbol{x}_i)_k$ and $\Sigma(\boldsymbol{x}_i)_k$ denote the $k$-th element of vectors $\mu(\boldsymbol{x}_i)$ and $\Sigma(\boldsymbol{x}_i)$. Considering the encoder and decoder processes, this $\tilde{\boldsymbol{x}}_i$ can be formulated as the following form:

$$\tilde{\boldsymbol{x}}_i = Dec(Enc(\boldsymbol{x}_i)) \quad (11)$$

In this proposed strategy, we used a linear neural network to construct the encoder of the VAE model by which the input $x_i$ is encoded into the latent representation $\boldsymbol{z}_i$. Thereafter, another linear neural network can be constructed for decoding $\boldsymbol{z}_i$ into $\tilde{\boldsymbol{x}}_i$.

Mathematically, for $Enc(\boldsymbol{x}_i)$ we have:

$$Enc(\boldsymbol{x}_i) = q_{\theta}(\boldsymbol{z}_i|\boldsymbol{x}_i) \sim \mathcal{N}\left(\tilde{\mu}(\boldsymbol{x}_i), \tilde{\Sigma}(\boldsymbol{x}_i)\right) \quad (12)$$

According to Eq. (2) and (3), $\tilde{\mu}(\boldsymbol{x}_i)$ and $\tilde{\Sigma}(\boldsymbol{x}_i)$ should therefore

be encoded through the following forms by the linear neural network:

$$\begin{cases} \tilde{\mu}(\boldsymbol{x}_i) = \boldsymbol{w}_2\sigma(\boldsymbol{x}_i\boldsymbol{w}_1) \\ \tilde{\Sigma}(\boldsymbol{x}_i) = \boldsymbol{w}_3\sigma(\boldsymbol{x}_i\boldsymbol{w}_1) \end{cases} \quad (13)$$

where $\boldsymbol{w}_1$ is the weight of the first layer in the encoder neural network. $\boldsymbol{w}_2$ and $\boldsymbol{w}_3$ are the weight of the second layer in the encoder neural network, $\sigma$ is the non-linear activation function. The latent representation $\boldsymbol{z}_i$ can then be expressed as:

$$\boldsymbol{z}_i = \tilde{\mu}(\boldsymbol{x}_i) + \tilde{\Sigma}(\boldsymbol{x}_i) \quad (14)$$

Similarly, for $Dec(\boldsymbol{z}_i)$ we have:

$$Dec(\boldsymbol{z}_i) = p(\boldsymbol{x}_i|\boldsymbol{z}_i) \sim \mathcal{N}\big(\mu(\boldsymbol{x}_i), \Sigma(\boldsymbol{x}_i)\big) \quad (15)$$

according to Eqs. (2) and (13), $\mu(\boldsymbol{x}_i)$ and $\Sigma(\boldsymbol{x}_i)$ should be decoded as the following form by the neural network:

$$\mu(\boldsymbol{x}_i) = \boldsymbol{w}_5\sigma(\boldsymbol{z}_i\boldsymbol{w}_4)$$
$$= \boldsymbol{w}_5\sigma((\boldsymbol{w}_2\sigma(\boldsymbol{x}_i\boldsymbol{w}_1) + \boldsymbol{w}_3\sigma(\boldsymbol{x}_i\boldsymbol{w}_1))\boldsymbol{w}_4) \quad (16)$$

$$\Sigma(\boldsymbol{x}_i) = \boldsymbol{w}_6\sigma(\boldsymbol{z}_i\boldsymbol{w}_4)$$
$$= \boldsymbol{w}_6\sigma((\boldsymbol{w}_2\sigma(\boldsymbol{x}_i\boldsymbol{w}_1) + \boldsymbol{w}_3\sigma(\boldsymbol{x}_i\boldsymbol{w}_1))\boldsymbol{w}_4) \quad (17)$$

where $\boldsymbol{w}_4$ is the weight of the first layer in the decoder neural network, $\boldsymbol{w}_5$ and $\boldsymbol{w}_6$ are the weight of second layer in the decoder neural network, the activation function $\sigma$ here is the same activation function as that of the encoder neural network.

The training target function $argmax_{\theta}ELBO(\theta)$ for our model was obtained by combining Eqs. (10), (11), and Eqs. (16) and (17):

$$\sum_{i,k}\left[\begin{array}{c}\left(x_i^k - Dec\left(Enc(x_i^k)\right)\right)^2 \\ +\frac{1}{2}\left(\begin{array}{c}\left(\boldsymbol{w}_6\sigma\left(\binom{\boldsymbol{w}_2\sigma(\boldsymbol{x}_i\boldsymbol{w}_1)}{+\boldsymbol{w}_3\sigma(\boldsymbol{x}_i\boldsymbol{w}_1)}\boldsymbol{w}_4\right)\right)_k \\ +\left(\left(\boldsymbol{w}_5\sigma\left(\binom{\boldsymbol{w}_2\sigma(\boldsymbol{x}_i\boldsymbol{w}_1)}{+\boldsymbol{w}_3\sigma(\boldsymbol{x}_i\boldsymbol{w}_1)}\boldsymbol{w}_4\right)\right)_k\right)^2 - 1 \\ -\log\boldsymbol{w}_6\sigma\left(\binom{\boldsymbol{w}_2\sigma(\boldsymbol{x}_i\boldsymbol{w}_1)}{+\boldsymbol{w}_3\sigma(\boldsymbol{x}_i\boldsymbol{w}_1)}\boldsymbol{w}_4\right)_k\end{array}\right)\end{array}\right] \quad (18)$$

The Variational Autoencoder (VAE) is a neural network used for generating artificial data that has a similar distribution to a given dataset. The training process involves iterating the maximization of ELBO: $argmax_\theta ELBO(\theta)$ (as expressed in Eq. 18) until the network produces appropriate weights $\boldsymbol{w_i}$ for the training dataset. During this process, the weights record information about the data distribution. This allows the VAE to generate artificial data (represented by $\tilde{\boldsymbol{z}}_i$) with a distribution similar to the original dataset $\boldsymbol{x}$ (as shown in Eqs. 16 and 17). This process ensures that the VAE can successfully generate artificial data that closely matches the original dataset. Once trained, the artificial feature generator can create artificial device features, and the artificial label generator can generate artificial device performance. For a better understanding of the training process and usage of the VAE model, refer to Fig. 2.

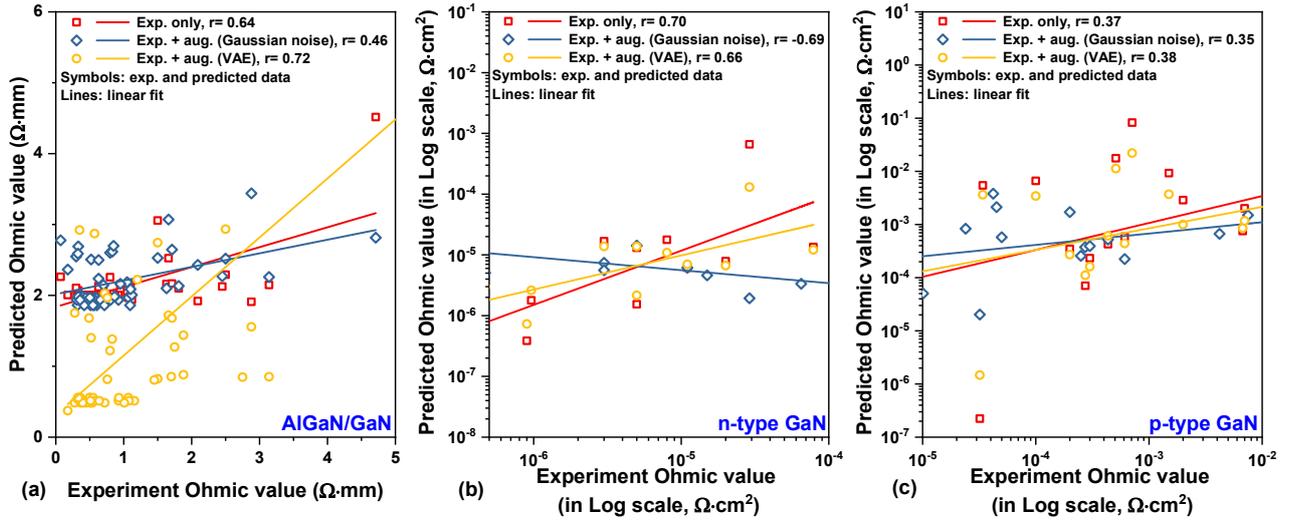

Fig. 5. Pearson r of the predicted and experimental ohmic resistance values of (a) AlGaN/GaN, (b) n-type GaN, and (c) p-type GaN substrates. The augmented data are 10 times larger than the experimental data. Note: fitting lines are from individual prediction process, which is not averaged. For the details of the mean values, kindly refer to Fig. 6.

#### 2) Artificial label Generator

The artificial label generator is a critical component of the data generation process, responsible for assigning device performances to the data generated by the artificial feature generator. Since the artificial and real features occupy the same space, the artificial performance data can be assumed to be in the same neighborhood as the real performance data. To generate performance values for artificial features, the nearest neighbor algorithm is employed to analyze the real performance data. This algorithm compares the artificial features with the real feature dataset to find the most similar real features, then assigns the corresponding performance value to the artificial features. In this way, the artificial label generator ensures that the generated artificial device performance closely matches the performance values of real devices. The detailed procedure for using the nearest neighbor algorithm to assign performance values to artificial features is explained below.

First, we calculate all the distances between artificial features and real features by an Euclidean metric [24]:

$$d(\dot{\boldsymbol{x}}_i, \boldsymbol{x}_j) = \sum_{k=1}^{M}(\dot{x}_{ik} - x_{jk})^2 \qquad (19)$$

where $d(\dot{\boldsymbol{x}}_i, \boldsymbol{x}_j)$ denotes the distance between $i$-th artificial feature $\dot{\boldsymbol{x}}_i$ and $j$-th real feature $\boldsymbol{x}_j$, $M$ is the dimension of $\boldsymbol{x}_j$.

Then, we introduce this distance into the performance data space. The nearest neighbor algorithm is applied to calculate the outputs of artificial features using the distance of features as follows:

$$\hat{y}_i = \frac{\sum_{r=1}^{S}\frac{1}{d(\dot{\boldsymbol{x}}_i, \boldsymbol{x}_r)}y_r}{\sum_{r=1}^{S}\frac{1}{d(\dot{\boldsymbol{x}}_i, \boldsymbol{x}_r)}} \qquad (20)$$

where $\hat{y}_i$ is the generated output of the $i$-th artificial feature $\dot{\boldsymbol{x}}_i$; $S$ denotes the top $S$ nearest neighbors to the $\dot{\boldsymbol{x}}_i$; and $y_r$ is the real output value of $r$-th nearest real feature. The artificial label generator first calculates the distance between an artificial vector and its nearest real data vectors.

Then the generator creates a corresponding artificial label by evaluating the real labels of its nearest neighbors (regarding the calculated distances) as expressed in Eq. (20). Note that the VAE models for AlGaN/GaN, n-GaN, and p-GaN data were trained separately in this paper, considering the intrinsic differences in the materials.

#### C. Methods for Verification

A DNN-based regression model is adopted in this paper to verify the artificial augmented data, (Fig. 1 sector IV). The model can be described in the following form:

$$\begin{cases} \boldsymbol{y} = \sigma(\boldsymbol{H}\boldsymbol{w_h} + \boldsymbol{b_h}) \\ \boldsymbol{H} = \sigma(\boldsymbol{X}\boldsymbol{w_i} + \boldsymbol{b_i}) \end{cases} \qquad (21)$$

where $\boldsymbol{y} \in \boldsymbol{R}^m$ are the measured values, $\boldsymbol{w_h} \in \boldsymbol{R}^n$ are the weight of the hidden layer, $\boldsymbol{b_h} \in \boldsymbol{R}^n$ is the bias of hidden layer, $\boldsymbol{w_i} \in \boldsymbol{R}^t$ are the weight of input layers, $\boldsymbol{b_i} \in \boldsymbol{R}^t$ is the bias of input layer, $\boldsymbol{X} \in \boldsymbol{R}^{m \times n}$ is the data matrix combined with the real samples and artificial samples, and $\sigma$ is the non-linear activation function.

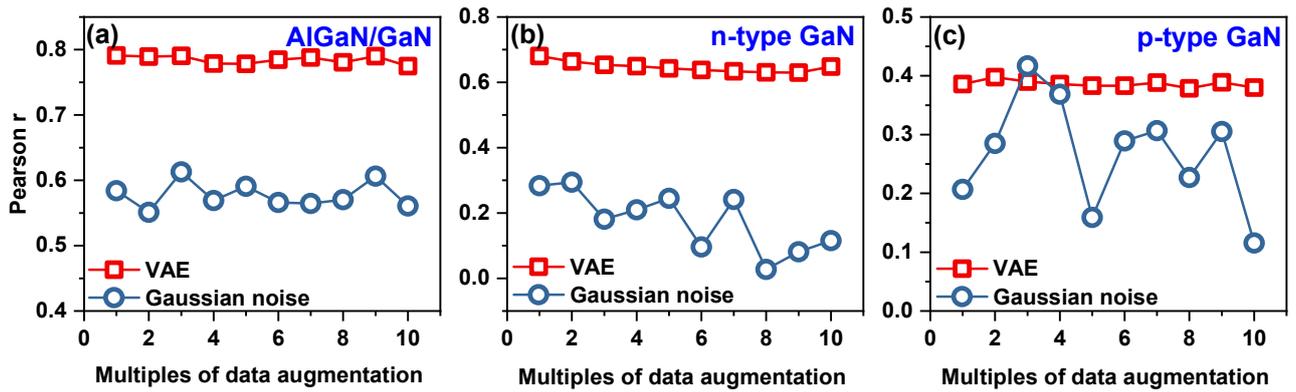

Fig. 6. The mean values of Pearson r (reflecting the correlation between the experimental data and the predicted data) using VAE-based model and noise-based model, with different augmentation scales for (a) AlGaN/GaN heterojunction, (b) n-type GaN, and (c) p-type GaN substrates. Note: each data point is averaged from five times predictions.

This model contains four layers including: an input layer for inputting the real features or artificial features; two hidden layers, and an output layer. The input layer has the same number of neurons as the length of those device features. The output layer contains only one neuron for predicting the device's electric performance. The two hidden layers that have more than 50 units are designed to fit the complex relationships between device features and their performance.

We used an experimental dataset of metal-semiconductor ohmic contact resistance, extracted from fabricated n-type GaN, p-type GaN, and AlGaN/GaN heterojunction devices (the dataset details can be found in our previous work [14]). The device structure and the process are represented in Fig. 3. The dataset includes resistance values and their corresponding fabrication recipes, such as metal layers, annealing temperature, annealing time, and annealing gas. This dataset is ideal for verifying our proposed self-augmentation model. This is because the dataset has few data points, a complicated fabrication process (also consuming significant foundry time), low fabrication recipe variation, and is difficult to simulate in TCAD.

Before training the DNN for the prediction task, we generated ten different scales of augmentations individually. These ranged from the same number of data points as the training dataset to ten times more. We then trained a DNN-based network for ohmic resistance prediction using a batch of random combinations of experimental and artificial data. We then tested its performance against the test data, which exclusively consisted of experimental data. Each DNN model was trained using augmented data at different scales, and we performed this process five times to ensure accuracy. For comparison, we used Gaussian noise augmentation as a control group and generated ten different scales of augmentations from a standard Gaussian distribution. Note that the DNN models for AlGaN/GaN, n-GaN, and p-GaN data were trained separately.

To evaluate the performance of the augmented dataset and the model, we measured the mean absolute error (MAE) of the prediction using a well-trained DNN-based ML model. To avoid bias towards the augmented data, we only used real experimental data for testing. There is, however, no standard index or figure-of-merit to evaluate the artificially generated

data. Therefore, we used three steps to evaluate our augmentation strategy:

1) *We mapped the augmented data into a lower dimension with experimental data for intuitive visualization.*

2) *We analyzed the Pearson r to evaluate the similarity between the prediction results from the augmented data and the experimental data.*

3) *We evaluated the MAE of the prediction task using augmented data.*

## III. RESULT AND DISCUSSION

### A. Augmented Data Visualization

Fig. 4 displays the kernel density plots that organize the distributions of the generated data with the real data. The Uniform Manifold Approximation and Projection (UMAP) algorithm was used to project the data from a high-dimensional parameter space into a 2-dimensional plane [25]. The artificially generated data is shown as a density color map, and the real data is shown as circles. It is observed that the artificially generated data is located in proximity to the real data. This indicates that it carries realistic information similar to the real data. Additionally, in Fig. 4, the high-concentration positions of the real data do not largely overlap the augmented data. This implies that the augmented data does not repeat the real data's pattern but (to some extent) compensates for the insufficient real data. The augmented data therefore can extend the occupied area in the data space, with a deliberate pattern, providing more comprehensive sampling points for machine learning (ML)-based tasks.

Furthermore, it can be observed that the kernel density of the augmented data is not condensed altogether in a small range but is dispersed over a large region. This observation suggests that the established augmentation model has successfully extracted the realistic patterns from the experimental data. It also suggests it has reasonably generated a more comprehensive artificial pattern containing sufficient additional realistic information.

### B. Pearson r of Augmented Data

To investigate and evaluate the proposed augmentation model, we plotted the ohmic resistance values obtained from predictions and experiments in Fig. 5. These predicted values were generated by the DNN-based prediction model. We discuss the evaluation of this model and its training process in the next subsection. Note that in Fig. 5, both the VAE-augmented and Gaussian noise augmented datasets have ten times more data points than the training data.

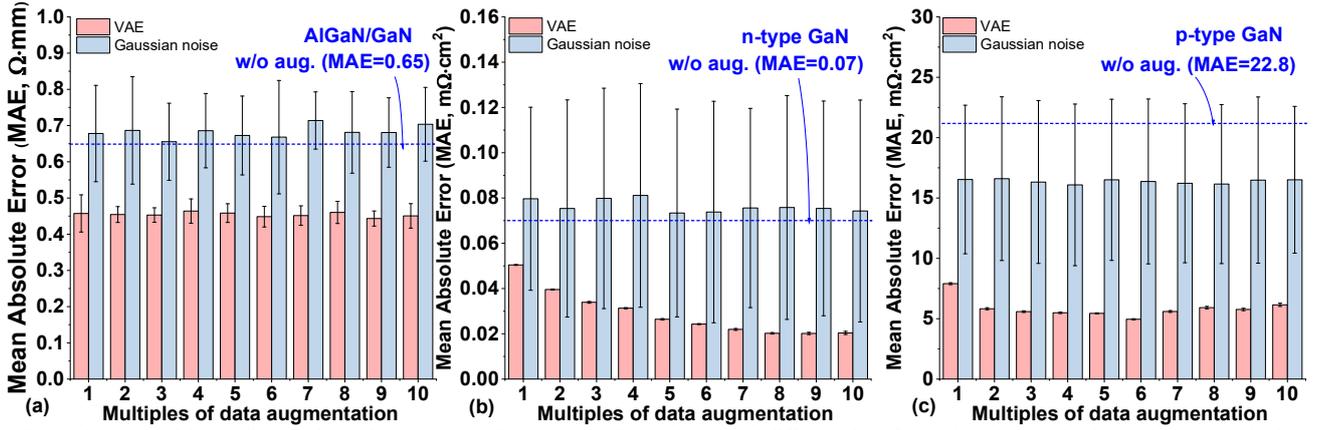

Fig. 7. The MAE of the prediction after 5 times testing, using pure experimental data, and different size of VAE-based and noise-based data, respectively, on (a) AlGaN/GaN, (b) n-type GaN, and (c) p-type GaN substrates. Each bar represents 5 repeated testing processes.

In Fig. 5, we observe that for all three types of substrates, the Pearson correlation coefficient of the model using VAE-augmented data is higher than that using Gaussian noise augmented data. The Gaussian noise can even result in a negative slope of the correlation sometimes (see n-GaN data). Moreover, the Pearson correlation coefficient of the augmented data in all three groups is similar to real experiment data, indicating that the generated augmentation data provides sufficient information similar to the real data.

The Pearson correlation coefficient of VAE-based data and noise-based data versus the augmentation scales (in multiples) is shown in Fig. 6 (mean values), where all but one point in the p-GaN group have higher r-index values for VAE-based data than for noise-based data. We observe a trend in Fig. 6 (b) and (c), where the Pearson correlation coefficient decreases with an increase in noise-based data points. This indicates that more noise-based data points lead to lower relativity, as the noise dilutes the data pool and fades the realistic information. This trend is not apparent in Fig. 6 (a), possibly due to the relatively sophisticated experimental data pool of the AlGaN/GaN group, where the augmented noise cannot significantly alter the data pattern. On the contrary to the noise-based data, the Pearson correlation coefficient of VAE-based data remains stable during ten times of multiplication, indicating that the proposed augmentation model does not inject any negative influence on the data pattern during augmentation. Thus, the data pool is not observed to be diluted.

### C. Mean Absolute Error of Prediction Task

Fig. 7 displays the mean absolute error (MAE) of the VAE-based and noise-based augmented data. The VAE-based model outperforms the noise-based model in all three device groups. For n-type GaN and p-type GaN, the MAE of the VAE-based model decreases initially and then flattens as the size of the augmented data increases. The MAE level of the noise-based model remains the same as the pure experimental data (without any augmentation, gray dashed lines) except for showing small fluctuations. This difference in behavior is attributed to the different levels of contribution of additional realistic information provided by VAE-based and noise-based data. The VAE-based data successfully exhibits the realistic data pattern, leading to a decrease in MAE. This pattern does however feature an accuracy limit (or systematic error), which eventually flattens the MAE at large augmentation scales. Increasing the amount of experimental data used to train the VAE would reduce this error further. Comparatively, the noise-based model can only contribute a random pattern to the DNN model, resulting in a shift in MAE. Notably, the error bar of the MAE of the noise-based data spans a huge range, whereas the VAE-based data provides a more confined MAE distribution in each multiple. This suggests that the augmented data from the VAE-based model is more patterned and less random than the noise-based data.

Interestingly, in Fig. 7 (a), the MAE of the VAE-based data does not show the same trend as its counterpart in Fig. 7 (b) and (c). Instead, it exhibits similar features to the noise-based data, although the mean MAE remains lower than the noise

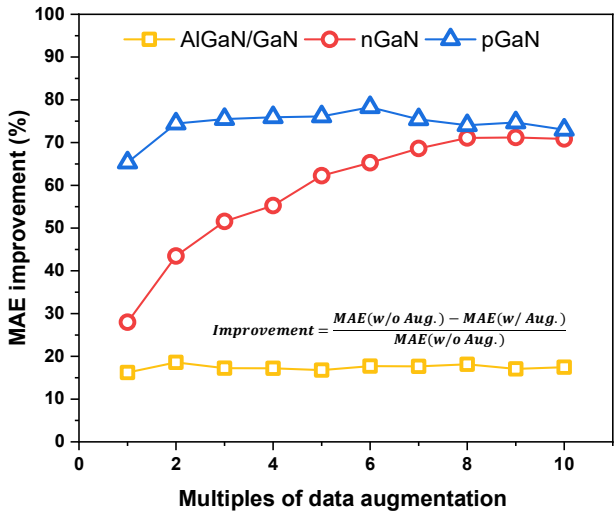

Fig. 8. MAE improvement in test process of different augmentation scale.

$$Improvement = \frac{MAE(w/o\ Aug.) - MAE(w/\ Aug.)}{MAE(w/o\ Aug.)}$$

group. The reason for this could be that the size of the experimental data in this group is larger than in the other groups, and, the augmentation model is not robust enough to extract sufficient patterns from such a large dataset. Alternatively, the augmentation model may only be able to extract partial information from the training data. This could especially be the case when the experimental data of this group is following several different patterns, as can be seen in Fig. 4 (a) where the circles are more dispersed than in the other two groups.

Fig. 8 provides more intuitive results of the MAE improvement provided by the augmentation. The augmented data significantly improves the prediction performance in the n-GaN and p-GaN groups, with an improvement of over 70 percent.

Although the proposed augmentation significantly improves the MAE of the modeling, the factors that contribute most to the enhancement are still not clear and require further exploration. This study suggests that the nature of the distribution of the experimental data may play a key role in this regard. Moreover, the source of the performance difference between AlGaN/GaN-type and other types of data could also be attributed to the features of the experimental data distribution, this warrants further investigation.

## IV. CONCLUSION

We have proposed and tested a VAE-based data self-augmentation strategy to relieve the contradiction between the accuracy and the insufficient training data in ML-based semiconductor device modeling. In this strategy, no additional TCAD simulation is required and only a few experimental data points are needed for functionality. The testing suggests that the established augmentation model could successfully extract realistic patterns from the experimental data, leading to a set of high-quality augmented data that was able to be seamlessly fed into the DNN model used. As a result, this strategy could significantly improve the performance of the DNN model, where a maximum of more than a 70% drop of MAE was obtained. We therefore believe this strategy could benefit the next-generation EDA simulations and modeling in the semiconductor industry.

## V. ACKNOWLEDGMENT

This work was partially supported by CSIRO's Impossible Without You program. Zeheng Wang acknowledges the support for the preliminary results of this work from Prof. Arne Laucht and the University of New South Wales.